\documentclass[10pt, a4paper]{article}
\usepackage{lrec}
\usepackage{multibib}
\newcites{languageresource}{Language Resources}
\usepackage{graphicx}
\usepackage{tabularx}
\usepackage{soul}
\usepackage{amsmath}
\usepackage{amsfonts}
\usepackage{amssymb}
\usepackage{subcaption}

\usepackage{epstopdf}
\usepackage[latin1]{inputenc}

\usepackage{hyperref}
\usepackage{xstring}
\usepackage{float}
\newcommand*\samethanks[1][\value{footnote}]{\footnotemark[#1]}
\DeclareMathOperator*{\argmax}{argmax}
\newcommand{\secref}[1]{\StrSubstitute{\getrefnumber{#1}}{.}{ }}

\title{Building a Word Segmenter for Sanskrit Overnight}

\name{Vikas Reddy\thanks{*The first two authors contributed equally}{*}\textsuperscript{1}, Amrith Krishna\samethanks[1]{*}\textsuperscript{2}, Vishnu Dutt Sharma\textsuperscript{3},\\
{\bf \large  Prateek Gupta\textsuperscript{4}, Vineeth M R\textsuperscript{5}, Pawan Goyal\textsuperscript{2}}}

\address{\textsuperscript{1}Dept. of Mining Engineering, \textsuperscript{2}Dept. of Computer Science and Engineering,
\textsuperscript{4}Dept. of Mathematics, \\\textsuperscript{5}Dept. of Electrical Engineering,         \textsuperscript{1,2,4,5}IIT Kharagpur \\ \textsuperscript{3}American Express India Pvt Ltd\\
         \{vikas.challaram, amrith\}@iitkgp.ac.in, pawang@cse.iitkgp.ernet.in\\
   }

\abstract{There is abundance of digitised texts available in Sanskrit. However, the word segmentation task in such texts are challenging due to the issue of \textit{Sandhi}. In Sandhi, words in a sentence often fuse together to form a single chunk of text, where the word delimiter vanishes and sounds at the word boundaries undergo transformations, which is also reflected in the written text. Here, we propose an approach that uses a deep sequence to sequence (seq2seq) model that takes only the sandhied string as the input and predicts the unsandhied string. The state of the art models  are linguistically involved and have external dependencies for the lexical and morphological analysis of the input. Our model can be trained ``overnight'' and be used for production. In spite of the knowledge lean approach, our system preforms better than the current state of the art by gaining a percentage increase of  16.79 \% than the current state of the art.\\ \newline  }

\begin{document}

\maketitleabstract

\section{Introduction}

Sanskrit had profound influence as the knowledge preserving language for centuries in India. The tradition of learning and teaching Sanskrit, though limited, still exists in India. There have been tremendous advancements in digitisation of ancient manuscripts in Sanskrit in the last decade. Numerous initiatives such as the Digital Corpus of Sanskrit\footnote{http://kjc-sv013.kjc.uni-heidelberg.de/dcs/}, GRETIL\footnote{http://gretil.sub.uni-goettingen.de/}, The Sanskrit Library\footnote{http://sanskritlibrary.org/} and others from the Sanskrit Linguistic and Computational Linguistic community is a fine example of such efforts \cite{goyal2012distributed,krishna-satuluri-goyal:2017:LaTeCH-CLfL}.   

The digitisation efforts have made the Sanskrit manuscripts easily available in the public domain. However, the accessibility of such digitised manuscripts is still limited. Numerous technical challenges in indexing and retrieval of such resources in a digital repository arise due to the linguistic peculiarities posed by the language. Word Segmentation in Sanskrit is an important yet non-trivial prerequisite for facilitating efficient processing of Sanskrit texts. Sanskrit has been primarily communicated orally. Due to its oral tradition, the phonemes in Sanskrit undergo euphonic assimilation in spoken format. This gets reflected in writing as well and leads to the phenomena of \textit{Sandhi} \cite{goyal2016design}. \textit{Sandhi} leads to phonetic transformations at word boundaries of a written chunk, and the sounds at the end of a word join together to form a single chunk of character sequence. This not only makes the word boundaries indistinguishable, but transformations occur to the characters at the word boundaries. The transformations can be deletion, insertion or substitution of one or more sounds at the word ends. There are about 281 sandhi rules, each denoting a unique combination of phonetic transformations, documented in the grammatical tradition of Sanskrit. The proximity between two  compatible sounds as per any one of the 281 rules is the sole criteria for sandhi.  The \textit{Sandhi} do not make any syntactic or semantic changes to the words involved. \textit{Sandhi} is an optional operation relied solely on the discretion of the writer.

While the \textit{Sandhi} formation is deterministic, the analysis of \textit{Sandhi} is non-deterministic and leads to high level of ambiguity.  For example, the chunk `{\sl gardabha\d{s}c{\=a}\'sva\'sca}' (the ass and the horse) has 625 possible phonetically and lexically valid splits~\cite{hellwigusing}. Now, the correct split relies on the semantic compatibility between the split words.

The word segmentation problem is a well studied problem across various languages where the segmentation is non-trivial. For languages such as Chinese and Japanese, where there is no explicit boundary markers between the words \cite{xue2003chinese}, numerous sequence labelling approaches have been proposed. In Sanskrit, it can be seen that the merging of word boundaries is the discretion of the writer. In this work, we propose a purely engineering based pipeline for segmentation of Sanskrit sentences. The word segmentation problem is a structured prediction problem and we propose a deep sequence to sequence (seq2seq) model to solve the task. We use an encoder-decoder framework where the \textit{sandhied} (unsegmented) and the \textit{unsandhied} (segmented) sequences are treated as the input at the encoder and the output at the decoder, respectively. We train the model so as to maximise the conditional probability of predicting the unsandhied sequence given its corresponding sandhied sequence \cite{cho-EtAl:2014:EMNLP2014}.  We propose a knowledge-lean data-centric approach for the segmentation task. Our approach will help to scale the segmentation process in comparison with the challenges posed by knowledge involved processes in the current systems \cite{krishna-satuluri-goyal:2017:LaTeCH-CLfL}.  We only use parallel segmented and unsegmented sentences during training. At run-time, we only require the input sentence.
Our model can literally be trained overnight.  
The best performing model of ours takes less than 12 hours to train in a `Titan X' 12 GB memory, 3584  GPU Cores system. Our title for the paper is inspired from the title for the work by \newcite{wang2015building}. As with the original paper, we want to emphasise on the ease with which our system can be used for training and at runtime, as it do not require any linguistically involved preprocessing. Such requirements often limit the scalability of a system and tediousness involved in the process limits the usability of a system.

Since Sanskrit is a resource scarce language, we use the \textit{sentencepiece} \cite{schuster2012japanese}, an unsupervised text tokeniser to obtain a new vocabulary for a corpus, that maximises the likelihood of the language model so learnt. We propose a pipeline for finding the semantically most valid segmented word-forms for a given sentence.  Our model uses multiple layers of LSTM cells with attention. Our model outperforms the current state of the art by 16.79 \%.

\section{Models for Word Segmentation in Sanskrit}
A number of methods have been proposed for word segmentation in Sanskrit.
\newcite{hellwigusing} treats the problem as a character level RNN sequence labelling task. The author, in addition to reporting sandhi splits to upto 155 cases, additionally categorises the rules to 5 different types. Since, the results reported by the author are not at word-level, as is the standard with word segmentation systems in general, a direct comparison with the other systems is not meaningful.  \newcite{mittal2010automatic} proposed a method based on Finite State Transducers by incorporating rules of \textit{sandhi}. The system generates all possible splits and then provides a ranking of various splits, based on probabilistic ranking inferred from a dataset of 25000 split points. Using the same dataset, \newcite{natarajan2011s3} proposed a sandhi splitter for Sanskrit. The method is an extension of Bayesian word segmentation approach by \newcite{goldwater2006contextual}. \newcite{krishna2016segmentation} is currently the state of the art in Sanskrit word segmentation. The system treats the problem as an iterative query expansion problem. Using a shallow parser for Sanskrit \cite{goyal2012distributed}, an input sentence is first converted to a graph of possible candidates and desirable nodes are iteratively selected using Path Constrained Random Walks \cite{lao2010relational}.  

To further catalyse the research in word segmentation for Sanskrit, \newcite{krishna-satuluri-goyal:2017:LaTeCH-CLfL} has released a dataset for the word segmentation task. The work releases a dataset of 119,000 sentences in Sanskrit along with the lexical and morphological analysis from a shallow parser.  
The work emphasises the need for not just predicting the inflected word form but also the prediction of the associated morphological information of the word. The additional information will be beneficial in further processing of Sanskrit texts, such as Dependency parsing or summarisation \cite{krishna-satuluri-goyal:2017:LaTeCH-CLfL}.
 So far, no system successfully predicts the morphological information of the words in addition to the final word form.  Though \newcite{krishna2016segmentation} has designed their system with this requirement in mind and outlined the possible extension of their system for the purpose, the system currently only predicts the final word-form.


\section{Method}
\label{method}

We use an encoder-decoder framework for tackling our segmentation problem, and propose a deep seq2seq model using LSTMs for our prediction task. Our model follows the architecture from \newcite{wu2016google}, originally proposed for neural machine translation. We consider the pair of \textit{sandhied} and \textit{unsandhied} sentences as source and target sentences, respectively. Following the insights from \newcite{sutskever2014sequence}, we reverse the sequence order at the input and we find that the reversal of the string leads to improvement in the results. We also use a deep architecture with 3 layers each at the encoder and decoder, as it is shown that deeper models perform better than shallow LSTM Models. We also experiment with models with and without attention and find that the model with attention leads to considerable improvement in performance of the system \cite{wu2016google}. Given the training set $\mathcal{S}$, our training objective is to maximise the log probability of the segmented sequences $T$ where the unsegmented sequences $S$ are given.
The training objective is to maximise \cite{sutskever2014sequence} 

$$
\frac{1}{|S|} \sum_{(T,S) \in S} log\  p(T|S)
$$

For a new sentence, we need to output a sequence $T'$ with maximum likelihood for the given input \cite{sutskever2014sequence}.

$$
T' = \argmax_T \ p(T|S)
$$

LSTMs are used both at the encoder and decoder. We use softmax layer at the decoder and perform greedy decoding to obtain the final prediction. The outputs are then passed to the loss function which calculates the log-perplexity over the data samples in the batch. We then update the parameters via backpropagation and
use Adam optimiser \cite{kingma2014adam} for our model.

\textbf{Vocabulary Enhancement for the model} - Sanskrit, being a resource poor language, the major challenge is to obtain enough data for the supervised task. While there are plenty of sandhied texts available for Sanskrit, it is hard to find parallel or unsandhied texts alone, as it is deterministic to get sandhied text from unsandhied texts. 

In our case we use 105,000 parallel strings from the Digital Corpus of Sanskrit as released in \newcite{krishna-satuluri-goyal:2017:LaTeCH-CLfL}. To handle the data sparsity, we adopt a purely engineering based approach for our model. Rather than relying on the real word boundaries, we use the `{\sl sentencepiece}' model, an unsupervised text tokeniser~\cite{schuster2012japanese} to obtain a new vocabulary for the corpus. The method was originally proposed for segmentation problem in Japanese and Korean speech recognition systems. In the method, a greedy approach is used to identify new word units from a corpus that maximises the likelihood of the language model so learnt \cite{schuster2012japanese}. 

\begin{figure}[!htb]
\centering
  \includegraphics[width=0.5\textwidth]{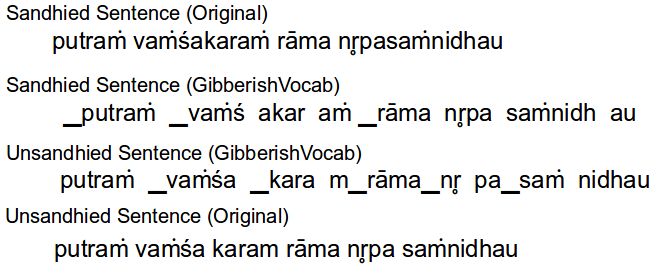}
\vspace{-1em}
  \caption{Sandhied and unsandhied sentence expressed in original writing and with the new learnt vocabulary `GibberishVocab'.}

  \label{archi}
\end{figure}

Figure \ref{archi} shows the instance of words learnt from the {\sl sentencepiece} model corresponding to the original input from the corpus. In the {\sl sentencepiece} model, the `space' in the original input is also treated as a character and is replaced with the special symbol `\_'. So `a\d{m}\_r{\=a}ma' is a word in our vocabulary, which originally is part of two words.

Our model is fed only the `words' from the new vocabulary, henceforth to be referred to as `\textit{GibberishVocab}'. Note that the decoder also outputs words from \textit{GibberishVocab}. The output from decoder is then converted to the original vocabulary for evaluating the outputs. This is trivially done by reclaiming the original `space' as the delimiter for the old vocabulary.

\section{Experiments}

\subsection{Dataset}

We used a dataset of 107,000 sentences from the Sanskrit Word Segmentation Dataset \cite{krishna-satuluri-goyal:2017:LaTeCH-CLfL}. The dataset is a subset of the Digital Corpus of Sanskrit. From the dataset we only use the input sentences and the ground truth inflected word-forms. We ignore all the other morphological and lemma information available in the dataset. 

\subsection{Baselines}

We compare the performance of our system with two other baseline systems.\\
\textbf{supervisedPCRW} - This is the current state of the art for word segmentation in Sanskrit. The method treats the problem as an iterative query expansion task. This is a linguistically involved approach, as at first a lexicon driven shallow parser is used to obtain all the phonetically valid segments for a given sentence. The sentence is then converted into a graph with the segments as the nodes. The edges are formed between every pair of nodes which can co-exist in a sentence and are not competing for the same input position. The edge weights are formed by weighted sum of random walks across typed paths. The authors use typed paths to obtain extended contexts about the word pairs from the candidate pool. The typed paths are designed with human supervision which is also linguistically involved.\\
\textbf{GraphCRF} - We use a structured prediction approach using graph based Conditional Random Fields, where we first obtain the possible candidate segments using the shallow parser and then convert the segments into a graph. For every node segment, we learn a word vector using fastText~\cite{bojanowski2016enriching}.  \\
\textbf{segSeq2Seq} - This is the proposed model as described in Section \ref{method} but without attention. \\
\textbf{attnSegSeq2Seq} - This is the proposed model as described in Section \ref{method} with attention.

We report all our results on a test data of 4,200 sentences which was not used in any part of the training. From the dataset we ignore about 7,500 sentences which are neither part of training nor the test set. We used 214,000 strings both from input and output strings in the training data to obtain the \textit{GibberishVocab} using {\sl sentencepiece} model. 

We use string-wise micro averaged precision, recall and F-Score to evaluate our model as is the standard with evaluating word segmentation models. We find that the default vocabulary size of 8,000 for the \textit{GibberishVocab} works best. Of the 8,000 `words', the encoder vocabulary size is 7,944 and the decoder vocabulary size is 7,464. This shows the high overlap in the vocabulary in \textit{GibberishVocab} at both input and output sides, in spite of the difference in phonetic transformations due to sandhi. Originally the training data contained 60,308 segmented words at the output side. By reducing the vocabulary size at decoder side to 7,464, we make the probability distribution (softmax) at the decoder layer denser. Even if we followed a linguistic approach there were 16,473 unique lemmas in the training dataset. 

\subsection{Training Procedure and Hyperparameters}

Our models have 3 layers at both the encoder and decoder. The models contain an embedding layer which is a trainable matrix with individual word vector having a size of 128. Our LSTM layers consist of 128 cells at both the encoder and decoder layers. We train the sentences in a batch size of 128 and keep the sequence length of each sequence to be 35. The initial learning rate was set at 0.001 and we trained our system for 80 epochs after which the network parameters converged. We used Adam optimiser with parameter values $\beta_1$,$\beta_2$ as 0.9 and 0.999, respectively. We use dropout in the hidden layers with different settings from 0.1 to 0.4 in step sizes of 0.1. We find that a dropout of 0.2 is the best performing configuration. Dropout helps to avoid over-fitting of data \cite{srivastava2014dropout}. Both the `segSeq2Seq' and `attnSegSeq2Seq' models  follow the same architecture and have the same hyperparameter settings and vary only on the attention mechanism.

\if{}
\textbf{TODO: this table can be removed if space is required.}
\begin{table}[!htb]
\centering
\begin{tabular}{|l|l|}
\hline
Vocabulary Size      & $8000$                                \\ \hline
Embedding Size  & 128 \\ \hline

Hidden Layer Size    & $128$                                 \\ \hline
Learning Rate        & $1 \times 10^ {-03}$ \\ \hline

Dropout Rate         & $0.095$                                 \\ \hline

Adam Optimizer ($\beta_1$,$\beta_2$,$\alpha$)         & $0.09,0.99,0.001$                                 \\ \hline

\end{tabular}
\caption{Hyper-parameters for the best performing configuration of the system.}
\label{hyper}
\end{table}
\fi

\begin{figure*}[t!]
   \centering
    \begin{subfigure}[t]{0.5\textwidth}
        \centering
        \includegraphics[height=2.2in]{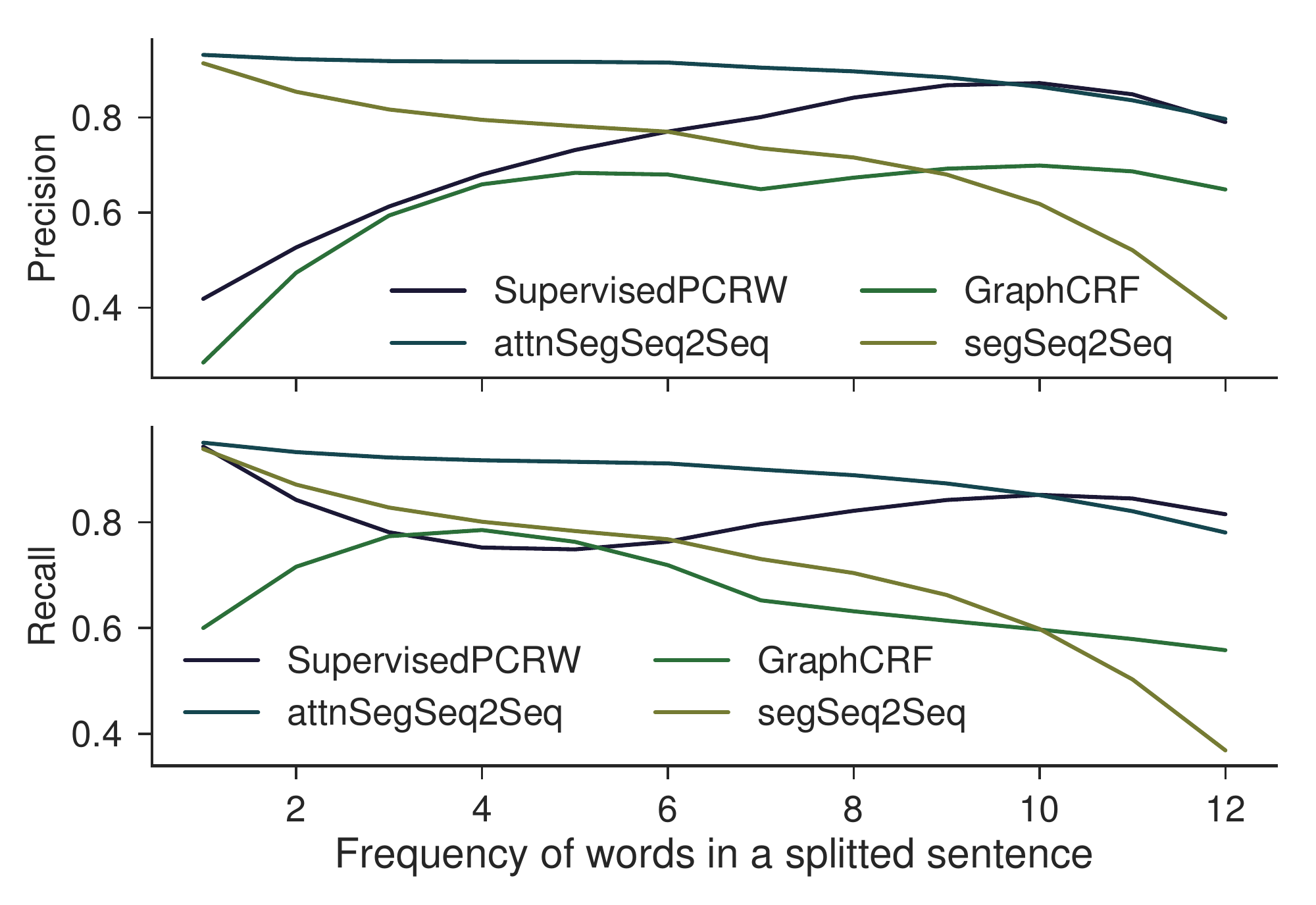}
        \caption{Precision and Recall for the competing systems grouped based on the count of words in each of the test
sentence.}
    \end{subfigure}%
    ~ 
    \begin{subfigure}[t]{0.5\textwidth}
        \centering
        \includegraphics[height=2.2in]{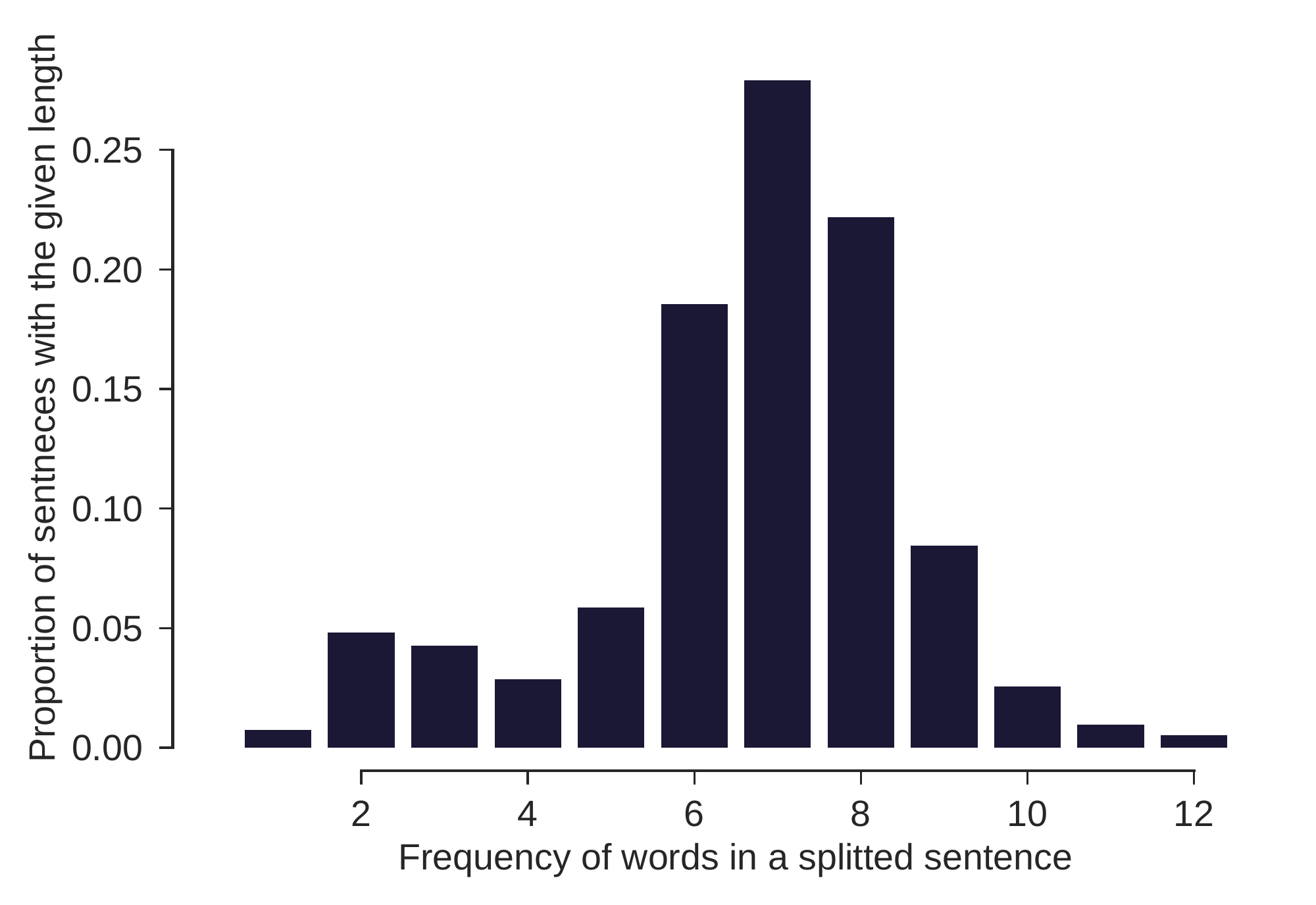}
        \caption{Distribution of strings in test dataset grouped based on the count of words in each sentence.}
    \end{subfigure}
    \vspace{-0.5em}
    \caption{Results on the test dataset. The sentences are grouped based on the count of words in the segmented sentences}
    \label{prec}
\vspace{-1.2em}
\end{figure*}

\subsection{Results}

\begin{table}[]
\centering
\begin{tabular}{|l|l|l|l|}
\hline
\textbf{Model} & \textbf{Precision} & \textbf{Recall} & \textbf{F-Score} \\ \hline
GraphCRF       &       65.20             &          66.50       &    65.84              \\ \hline
SupervisedPCRW &     76.30               &   79.47              & 77.85                  \\ \hline
segSeq2Seq     & 73.44             & 73.04           & 73.24           \\ \hline
attnsegSeq2Seq     & 90.77             & 90.3           & 90.53           \\ \hline

\end{tabular}
\caption{Micro-averaged Precision, Recall and F-Score for the competing systems on the test dataset of 4200 strings.}
\label{results}
\end{table}

Table \ref{results} shows the performance of the competing systems. We can find that the system `attnSegSeq2Seq' outperforms the current state of the art with a percent increase of 16.29 \% in F-Score. The model `segSeq2Seq' falls short of the current state of the art with a percent decrease of  6.29 \% in F-Score. It needs to be noted that the systems `attnSegSeq2Seq' and `segSeq2Seq' are exactly the same architectures other than the addition of attention in the former. But there is a percentage increase of 23.61 \% for both the systems. One probable reason for this is due to the free word order nature of sentences in Sanskrit. Since there are multiple permutations of words in a sentence which are valid syntactically and convey the same semantic meaning, the entire input context is required to understand the meaning of a sentence for any distributional semantic model. \\

Figure \ref{prec} shows the results of the competing systems on strings of different lengths in terms of words in the sentence. This should not be confused with sequence length. Here, we mean the `word' as per the original vocabulary and is common for all the competing systems. For all the strings with up to 10 words, our system `attnSegSeq2Seq' consistently outperforms all the systems in terms of both precision and recall. The current state of the art performs slightly better than our system, for sentences with more than 10 words. It needs to be noted that the average length of a string in the Digital Corpus of Sanskrit is 6.7 \cite{krishna2016segmentation}. The proportion of sentences with more than 10 words in our dataset is less than 1 \%. The test dataset has slightly more than 4 \% sentences with 10 or more words. The `segSeq2Seq' model performs better than the state of the art for both Precision and Recall for strings with less than or equal to 6 words. Figure \ref{prec}a shows the proportion of sentences in the test data based on the frequency of words in it. Figure \ref{prec}b shows the proportion of strings in the test dataset based on the number of words in the strings. Our systems attnSegSeq2Seq takes overall 11 hours 40 minutes and  for 80 epochs in a `Titan X' 12GB GPU memory, 3584 GPU Cores, 62GB RAM and Intel Xeon CPU E5-2620 2.40GHz system. For segSeq2Seq it takes 7 hours for the same setting.

\section{Discussion}

The purpose of our proposed model is purely to identify the word splits and correctness of the inflected word forms from a sandhied string. 
The word-level indexing in retrieval systems is often affected by phonetic transformations in words due to \textit{sandhi}. For example, the term `\textit{parame\'{s}vara\d{h}}' is split as `\textit{parama}' (ultimate) and `\textit{{\=i}\'svara\d{h}}' (god). Now, a search for instances of the word `\textit{{\=i}\'svara\d{h}}' might lead to  missing search results without proper indexing. String matching approaches often result in low precision results. Using a lexicon driven system might alleviate the said issues, but can lead to possible splits which are not semantically compatible. For \textit{parame\'{s}vara\d{h}}', it can be split as `\textit{parama}' (ultimate), `\textit{\'sva}' (dog) and `\textit{ra\d{h}}' (to facilitate). Though this is not semantically meaningful it is lexically valid. Such tools are put to use by some of the existing systems \cite{krishna2016segmentation,mittal2010automatic} to obtain additional morphological or syntactic information about the sentences. This limits the scalability of those systems, as they cannot handle out of vocabulary words. Scalability of such systems is further restricted as the sentences often need to undergo linguistically involved preprocessing steps that lead to human in the loop processing. The systems by  \newcite{krishna2016segmentation} and \newcite{krishna-satuluri-goyal:2017:LaTeCH-CLfL} 
assume that the parser by \newcite{goyal2012distributed}, identifies all the possible candidate chunks. 

Our proposed model is built with precisely one purpose in mind, which is to predict the final word-forms in a given sequence. \newcite{krishna-satuluri-goyal:2017:LaTeCH-CLfL} states that it is desirable to predict the morphological information of a word from along with the final word-form as the information will be helpful in further processing of Sanskrit. The segmentation task is seen as a means and not an end itself. Here, we overlook this aspect and see the segmentation task as an end in itself. So we achieve scalability at the cost of missing out on providing valuable linguistic information. Models that use linguistic resources are at an advantage here. Those systems such as \newcite{krishna2016segmentation} can be used to identify the morphological tags of the system as they currently store the morphological information of predicted candidates, but do not use them for evaluation as of now. Currently, no system exists that performs the prediction of wordform and morphological information jointly for Sanskrit.  In our case, since we learn a new vocabulary altogether, the real word boundaries are opaque to the system. The decoder predicts from its own vocabulary. But predicting morphological information requires the knowledge of exact word boundaries. This should be seen as a multitask learning set up. One possible solution is to learn `\textit{GibberishVocab}' on the set of words rather than sentences. But this leads to increased vocabulary at decoder which is not beneficial, given the scarcity of the data we have. Given the importance of morphological segmentation in morphologically rich languages such as Hebrew and Arabic \cite{seeker2015graph}, the same applies to the morphologically rich Sanskrit as well \cite{krishna-satuluri-goyal:2017:LaTeCH-CLfL}. But,  we leave this work for future.

\section{Conclusion}

In this work we presented a model for word segmentation in Sanskrit using a purely engineering based appraoch. Our model with attention outperforms the current state of the art \cite{krishna2016segmentation}. 
Since, we tackle the problem with a non-linguistic approach, we hope to extend the work to other Indic languages as well where sandhi is prevalent such as Hindi, Marathi, Malayalam, Telugu etc. Since we find that the inclusion of attention is highly beneficial in improving the performance of the system, we intend to experiment with recent advances in the encoder-decoder architectures, such as \newcite{vaswani2017attention} and  \newcite{gehring2017convolutional}, where different novel approaches in using attention are experimented with. Our experiments in line with the measures reported in \newcite{krishna2016segmentation} show that our system performs robustly across strings of varying word size.  

\section*{Code and Dataset}
All our working code can be downloaded at \url{https://github.com/cvikasreddy/skt }. The dataset for training can be downloaded at \url{https://zenodo.org/record/803508#.WTuKbSa9UUs}

\if{}
\subsection{General Instructions for the Submitted Abstract}

Each submitted abstract should be between \ul{a minimum of three and
a maximum of four pages including figures}.

\section{Paper}

Each manuscript should be submitted on white A4 paper. The fully
justified text should be formatted in two parallel columns, each 8.25 cm wide,
and separated by a space of 0.63 cm. Left, right, and bottom margins should be
1.9 cm. and the top margin 2.5 cm. The font for the main body of the text should
be Times New Roman 10 with interlinear spacing of 12 pt.  Articles must be
between 4 and 8 pages in length, regardless of the mode of presentation (oral
or poster).

\subsection{General Instructions for the Final Paper}

Each paper is allocated between \ul{a minimum of four and a maximum of
eight pages including figures}. The unprotected PDF files will appear in the
on-line proceedings directly as received. Do not print the page number.

\section{Page Numbering}

\textbf{Please do not include page numbers in your article.} The definitive page
numbering of articles published in the proceedings will be decided by the
organising committee.

\section{Headings / Level 1 Headings}

Headings should be capitalised in the same way as the main title, and centred
within the column. The font used is Times New Roman 12 bold. There should
also be a space of 12 pt between the title and the preceding section, and
a space of 3 pt between the title and the text following it.

\subsection{Level 2 Headings}

The format for level 2 headings is the same as for level 1 Headings, with the
font Times New Roman 11, and the heading is justified to the left of the column.
There should also be a space of 6 pt between the title and the preceding
section, and a space of 3 pt between the title and the text following it.

\subsubsection{Level 3 Headings}

The format for level 3 headings is the same as for level 2 headings, except that
the font is Times New Roman 10, and there should be no space left between the
heading and the text. There should also be a space of 6 pt between the title and
the preceding section, and a space of 3 pt between the title and the text
following it.

%

\section{Citing References in the Text}

\subsection{Bibliographical References}

All bibliographical references within the text should be put in between
parentheses with the author's surname followed by a comma before the date
of publication \cite{Martin-90}. If the sentence already includes the author's
name, then it is only necessary to put the date in parentheses:
\newcite{Martin-90}. When several authors are cited, those references should be
separated with a semicolon: \cite{Martin-90,CastorPollux-92}. When the reference
has more than three authors, only cite the name of the first author followed by
``et al.'' (e.g. \cite{Superman-Batman-Catwoman-Spiderman-00}).

\subsection{Language Resource References}

\subsubsection{When Citing Language Resources}

When citing language resources, we recommend to proceed in the same way to
bibliographical references, except that, in order to make them appear in
a separate section, you need to use the \texttt{\\citelanguageresource} tag.
Thus, a language resource should be cited as \citelanguageresource{speecon}.

\subsubsection{When Not Citing Any Language Resource}

When no language resource needs to be cited in the paper, you need to comment
out a few lines in the \texttt{.tex} file:

\begin{verbatim}
% \usepackage{multibib}
% \newcites{languageresource}{}
% \section{Language Resource References}
% \bibliographystylelanguageresource
%   {lrec}
% \bibliographylanguageresource{xample}
\end{verbatim}

\section{Figures \& Tables}
\subsection{Figures}

All figures should be centred and clearly distinguishable. They should never be
drawn by hand, and the lines must be very dark in order to ensure a high-quality
printed version. Figures should be numbered in the text, and have a caption in
Times New Roman 10 pt underneath. A space must be left between each figure and
its respective caption. 

Example of a figure enclosed in a box:

\begin{figure}[!h]
\begin{center}
\includegraphics[scale=0.5]{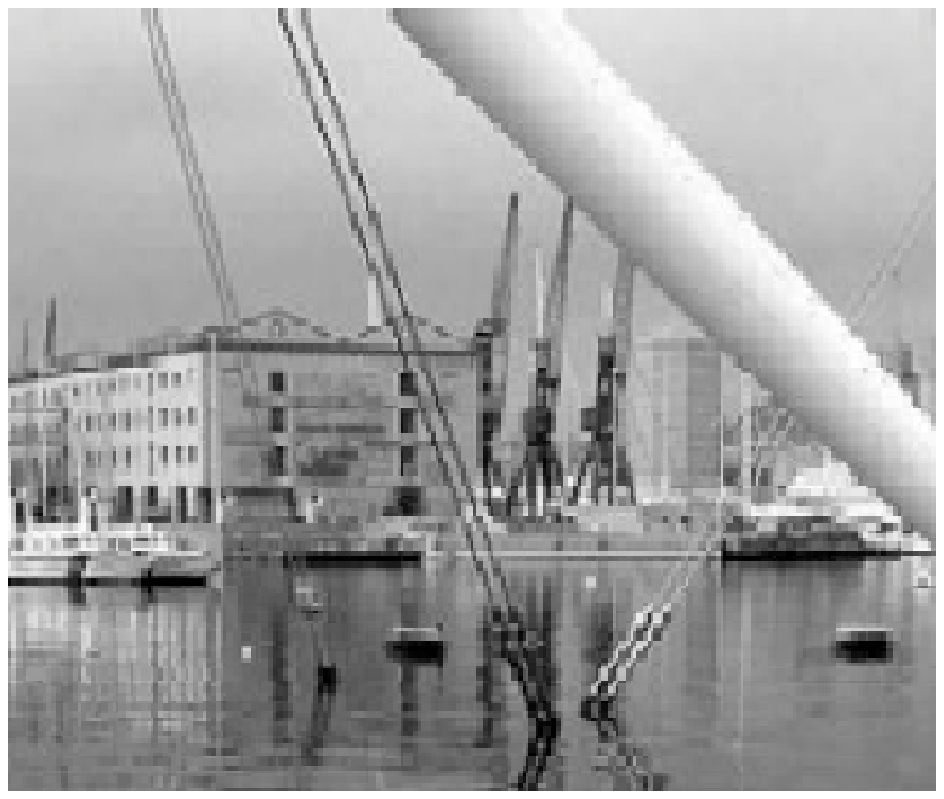} 
\caption{The caption of the figure.}
\label{fig.1}
\end{center}
\end{figure}

Figure and caption should always appear together on the same page. Large figures
can be centred, using a full page.

\subsection{Tables}

The instructions for tables are the same as for figures.
%
\begin{table}[!h]
\begin{center}
\begin{tabularx}{\columnwidth}{|l|X|}

      \hline
      Level&Tools\\
      \hline
      Morphology & Pitrat Analyser\\
      \hline
      Syntax & LFG Analyser (C-Structure)\\
      \hline
     Semantics & LFG F-Structures + Sowa's\\
     & Conceptual Graphs\\
      \hline

\end{tabularx}
\caption{The caption of the table}
 \end{center}
\end{table}

%
%
%
%
%

\section{Footnotes}

Footnotes are indicated within the text by a number in
superscript\footnote{Footnotes should be in Times New Roman 9 pt, and appear at
the bottom of the same page as their corresponding number. Footnotes should also
be separated from the rest of the text by a horizontal line 5 cm long.}.

\section{Copyrights}

The Lan\-gua\-ge Re\-sour\-ce and Evalua\-tion Con\-fe\-rence (LREC)
proceedings are published by the European Language Resources Association (ELRA).
They are available online from the conference website.

ELRA's policy is to acquire copyright for all LREC contributions. In assigning
your copyright, you are not forfeiting your right to use your contribution
elsewhere. This you may do without seeking permission and is subject only to
normal acknowledgement to the LREC proceedings. The LREC 2018 Proceedings are
licensed under CC-BY-NC, the Creative Commons Attribution-Non-Commercial 4.0
International License.

\section{Conclusion}

Your submission of a finalised contribution for inclusion in the LREC
proceedings automatically assigns the above-mentioned copyright to ELRA.

\section{Acknowledgements}

Place all acknowledgements (including those concerning research grants and
funding) in a separate section at the end of the article.

\section{Providing References}

\subsection{Bibliographical References}
Bibliographical references should be listed in alphabetical order at the
end of the article. The title of the section, ``Bibliographical References'',
should be a level 1 heading. The first line of each bibliographical reference
should be justified to the left of the column, and the rest of the entry should
be indented by 0.35 cm.

The examples provided in Section \secref{main:ref} (some of which are fictitious
references) illustrate the basic format required for articles in conference
proceedings, books, journal articles, PhD theses, and chapters of books.

\subsection{Language Resource References}

Language resource references should be listed in alphabetical order at the end
of the article, in the \textbf{Language Resource References} section, placed after
the \textbf{Bibliographical References} section. The title of the ``Language Resource
References'' section, should be a level 1 heading. The first line of each
language resource reference should be justified to the left of the column, and
the rest of the entry should be indented by 0.35 cm. The example in Section 
\secref{lr:ref} illustrates the basic format required for language resources.

In order to be able to cite a language resource, it must be added to
the \texttt{.bib} file first, as a \texttt{@LanguageResource} item type, which
contains the following fields:

\begin{itemize}
    \item{\texttt{author}: the builder of the resource}
    \item{\texttt{title}: the name of the resource}
    \item{\texttt{publisher}: the publisher of the resource (project,
          organisation etc)}
    \item{\texttt{year}: year of the resource release}
    \item{\texttt{series}: more general resource set this language resource
          belongs to}
    \item{\texttt{edition}: version of the resource}
    \item{\texttt{islrn}: the International Standard Language Resource Number
          (ISLRN) of the resource\footnote{The ISLRN number is available from
          \texttt{http://islrn.org}}} 
\end{itemize}

If you want the full resource author name to appear in the citation, the
language resource author name should be protected by enclosing it between
\texttt{\{...\}}, as shown in the model \texttt{.bib} file.

\vspace{.3\baselineskip}

\section*{Appendix: How to Produce the \texttt{.pdf} Version}

In order to generate a PDF file out of the LaTeX file herein, when citing
language resources, the following steps need to be performed:

\begin{itemize}
    \item{Compile the \texttt{.tex} file once}
    \item{Invoke \texttt{bibtex} on the eponymous \texttt{.aux} file}
    \item{Invoke \texttt{bibtex} on the \texttt{languageresources.aux} file}
    \item{Compile the \texttt{.tex} file twice}
\end{itemize}

\fi{}

\section{Bibliographical References}
\label{main:ref}

\bibliographystyle{lrec}
\bibliography{xample}

\label{lr:ref}

\bibliographystylelanguageresource{lrec}
\bibliographylanguageresource{xample}

\end{document}